\documentclass[utf8]{frontiersSCNS} 

\usepackage{url,hyperref,lineno,microtype,subcaption}
\usepackage[onehalfspacing]{setspace}
\usepackage[linesnumbered,lined,commentsnumbered]{algorithm2e}

\def\keyFont{\fontsize{8}{11}\helveticabold }
\def\firstAuthorLast{Sengupta {et~al.}} 
\def\Authors{Abhronil Sengupta\,$^{1,*}$, Yuting Ye\,$^{2}$, Robert Wang\,$^{2}$, Chiao Liu\,$^{2}$ and Kaushik Roy\,$^{1}$}
\def\Address{$^{1}$Department of Electrical and Computer Engineering, Purdue University, West Lafayette, IN, USA \\
$^{2}$Facebook Reality Labs, Facebook Research, Redmond, WA, USA  }
\def\corrAuthor{Abhronil Sengupta}

\def\corrEmail{asengup@purdue.edu}

\begin{document}
\onecolumn
\firstpage{1}

\title[Going Deeper in Spiking Neural Networks]{Going Deeper in Spiking Neural Networks: VGG and Residual Architectures} 

\author[\firstAuthorLast ]{\Authors} 
\address{} 
\correspondance{} 

\extraAuth{}

\maketitle

\begin{abstract}

\section{}
Over the past few years, Spiking Neural Networks (SNNs) have become popular as a possible pathway to enable low-power event-driven neuromorphic hardware. However, their application in machine learning have largely been limited to very shallow neural network architectures for simple problems. In this paper, we propose a novel algorithmic technique for generating an SNN with a deep architecture, and demonstrate its effectiveness on complex visual recognition problems such as CIFAR-10 and ImageNet. Our technique applies to both VGG and Residual network architectures, with significantly better accuracy than the state-of-the-art. Finally, we present analysis of the sparse event-driven computations to demonstrate reduced hardware overhead when operating in the spiking domain.

\tiny
 \keyFont{ \section{Keywords:} Spiking Neural Networks, Event-Driven Neural Networks, Sparsity, Neuromorphic Computing, Visual Recognition} 
\end{abstract}

\section{Introduction}

Spiking Neural Networks (SNNs) are a significant shift from the standard way of operation of Artificial Neural Networks (\cite {farabet2012comparison}). Most of the success of deep learning models of neural networks in complex pattern recognition tasks are based on neural units that receive, process and transmit analog information. Such Analog Neural Networks (ANNs), however, disregard the fact that the biological neurons in the brain (the computing framework after which it is inspired) processes binary spike-based information. Driven by this observation, the past few years have witnessed significant progress in the modeling and formulation of training schemes for SNNs as a new computing paradigm that can potentially replace ANNs as the next generation of Neural Networks.
In addition to the fact that SNNs are inherently more biologically plausible, they offer the prospect of event-driven hardware operation. Spiking Neurons process input information only on the receipt of incoming binary spike signals.
Given a sparsely-distributed input spike train, the hardware overhead (power consumption) for such a spike or event-based hardware would be significantly reduced since large sections of the network that are not driven by incoming spikes can be power-gated (\cite{chen1998estimation}). However, the vast majority of research on SNNs have been limited to very simple and shallow network architectures on relatively simple digit recognition datasets like MNIST (\cite{lecun1998gradient}) while only few works report their performance on more complex standard vision datasets like CIFAR-10 (\cite{krizhevsky2009learning}) and ImageNet (\cite{russakovsky2015imagenet}).
The main reason behind their limited performance stems from the fact that SNNs are a significant shift from the operation of ANNs due to their temporal information processing capability. This has necessitated a rethinking of training mechanisms for SNNs.

\section{Related Work}
Broadly, there are two main categories for training SNNs - supervised and unsupervised. Although unsupervised learning mechanisms like Spike-Timing Dependent Plasticity (STDP) are attractive for the implementation of low-power on-chip local learning, their performance is still outperformed by supervised networks on even simple digit recognition platforms like the MNIST dataset (\cite{diehl2015unsupervised}). Driven by this fact, a particular category of supervised SNN learning algorithms attempts to train ANNs using standard training schemes like backpropagation (to leverage the superior performance of standard training techniques for ANNs) and subsequently convert to event-driven SNNs for network operation (\cite{diehl2015fast,cao2015spiking,zhao2015feedforward,perez2013mapping}). This can be particularly appealing for NN implementations in low-power neuromorphic hardware specialized for SNNs (\cite{merolla2014million,akopyan2015truenorth}) or interfacing with silicon cochleas or event-driven sensors (\cite{posch2014retinomorphic,posch2011qvga}). Our work falls in this category and is based on the ANN-SNN conversion scheme proposed by authors in Ref. (\cite{diehl2015fast}). However, while prior work considers the ANN operation only during the conversion process, we show that considering the actual SNN operation during the conversion step is crucial for achieving minimal loss in classification accuracy. To that effect, we propose a novel weight-normalization technique that ensures that the actual SNN operation is in the loop during the conversion phase.  
Note that this work tries to exploit neural activation sparsity by converting networks to the spiking domain for power-efficient hardware implementation and are complementary to efforts aimed at exploring sparsity in synaptic connections (\cite{han2015deep}).

\section{Main Contributions}
The specific contributions of our work are as follows:

(i) As will be explained in later sections, there are various architectural constraints involved for training ANNs that can be converted to SNNs in a near-lossless manner. Hence, it is unclear whether the proposed techniques would scale to larger and deeper architectures for more complicated tasks. \textbf{We provide proof of concept experiments that deep SNNs (extending from 16 to 34 layers) can provide competitive accuracies over complex datasets like CIFAR-10 and ImageNet.}

(ii) \textbf{We propose a new ANN-SNN conversion technique that statistically outperforms state-of-the-art techniques.} We report a classification error of \textbf{8.45\%} on the CIFAR-10 dataset which is the best-performing result reported for any SNN network, till date. For the first time, we report an SNN performance on the entire ImageNet 2012 validation set. We achieve a \textbf{30.04\%} top-1 error rate and \textbf{10.99\%} top-5 error rate for VGG-16 architectures.

(iii) We explore Residual Network (ResNet) architectures as a potential pathway to enable deeper SNNs. We present insights and design constraints that are required to ensure ANN-SNN conversion for ResNets. We report a classification error of \textbf{12.54\%} on the CIFAR-10 dataset and a \textbf{34.53\%} top-1 error rate and \textbf{13.67\%} top-5 error rate on the ImageNet validation set. \textbf{This is the first work that attempts to explore SNNs with residual network architectures.}

(iv) \textbf{We demonstrate that SNN network sparsity significantly increases as the network depth increases.} This further motivates the exploration of converting ANNs to SNNs for event-driven operation to reduce compute overhead.

\section{Preliminaries}
\label{Preliminaries}
\subsection{Input and Output Representation}
\begin{figure}[!t]
\centering
\includegraphics[width=4.5in]{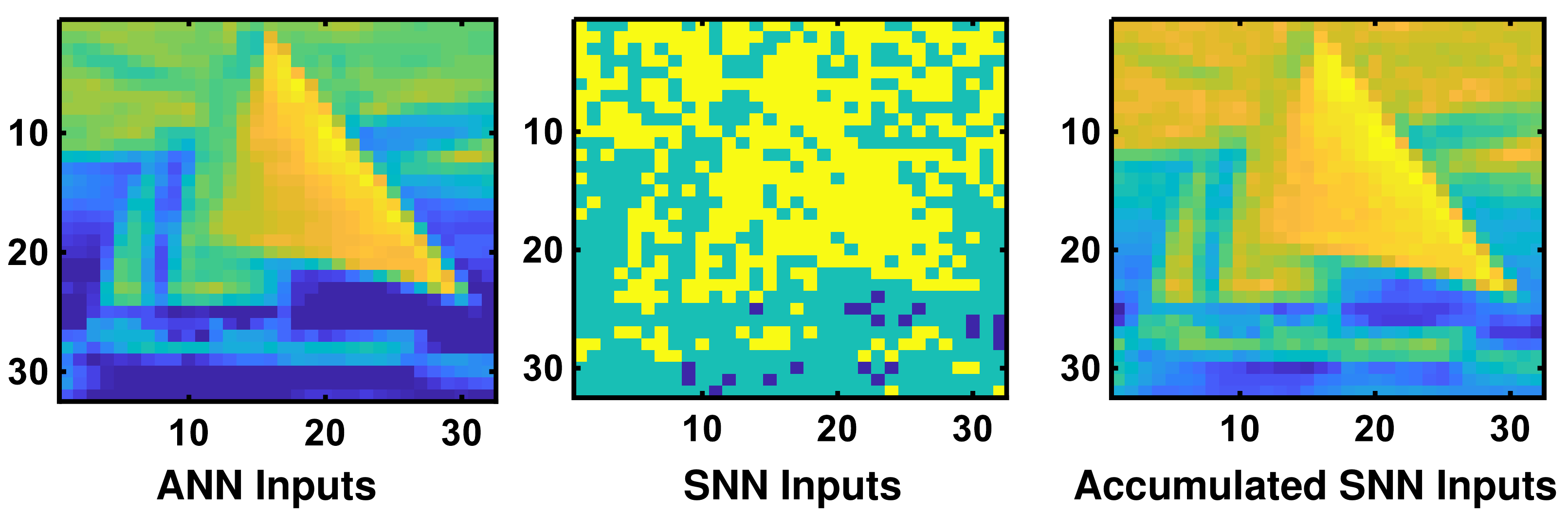}
\caption{The extreme left panel depicts a particular input image from the CIFAR-10 dataset with per pixel mean (over the training set) subtracted that is provided as input to the original ANN. The middle panel represents a particular instance of the Poisson spike train generated from the analog input image. The accumulated events provided to the SNN over $1000$ timesteps is depicted in the extreme right panel. This justifies the fact that the input image is being rate encoded over time for SNN operation.}
\label{fig1_a}
\end{figure}
The main difference between ANN and SNN operation is the notion of time. While ANN inputs are static, SNNs operate based on dynamic binary spiking inputs as a function of time. 
The neural nodes also receive and transmit binary spike input signals in SNNs, unlike in ANNs, where the inputs and outputs of the neural nodes are analog values. 
In this work, we consider a rate-encoded network operation where the average number of spikes transmitted as input to the network over a large enough time window is approximately proportional to the magnitude of the original ANN inputs (pixel intensity in this case). The duration of the time window is dictated by the desired network performance (for instance, classification accuracy) at the output layer of the network. A Poisson event-generation process is used to produce the input spike train to the network. Every time-step of SNN operation is associated with the generation of a random number whose value is compared against the magnitude of the corresponding input. A spike event is triggered if the generated random number is less than the value of the corresponding pixel intensity. This process ensures that the average number of input spikes in the SNN is proportional to the magnitude of the corresponding ANN inputs and is typically used to simulate an SNN for recognition tasks based on datasets for static images (\cite{diehl2015fast}). Fig. \ref{fig1_a} depicts a particular timed-snapshot of the input spikes transmitted to the SNN for a particular image from the CIFAR-10 dataset.  Note that since we are considering per pixel mean subtracted images, the input layer receives spikes whose rate is proportional to the input magnitude with sign equal to the input sign. However, for subsequent layers all spikes are positive in sign since there are generated by spiking neurons in the network. SNN operation of such networks are “pseudo-simultaneous”, i.e. a particular layer operates immediately on the incoming spikes from the previous layer and does not have to wait for multiple time-steps for information from the previous layer neurons to get accumulated. Given a Poisson-generated spike train being fed to the network, spikes will be produced at the network outputs. Inference is based on the cumulative spike count of neurons at the output layer of the network over a given time-window.

\subsection{ANN and SNN Neural Operation}
ANN to SNN conversion schemes usually consider Rectified Linear Unit (ReLU) as the ANN activation function. For a neuron receiving inputs $x_i$ through synaptic weights $w_i$, the ReLU neuron output $y$ is given by,
\begin{equation}
y = max \left( 0,\sum_{i} w_i.x_i \right)
\end{equation}
Although ReLU neurons are typically used in a large number of machine learning tasks at present, the main reason behind their usage for ANN-SNN conversion schemes is that they bear functional equivalence to an Integrate-Fire (IF) Spiking Neuron without any leak and refractory period (\cite{cao2015spiking,diehl2015fast}). 
Note that this is a particular type of Spiking Neuron model (\cite{izhikevich2003simple}). Let us consider the ANN inputs $x_i$ encoded in time as a spike train $\mathbb{X}_i(t)$, where the average value of $\mathbb{X}_i(t)$, $\mathbb{E}[\mathbb{X}_i(t)] \propto x_i$ (for the rate encoding network being considered in this work). The IF Spiking Neuron keeps track of its membrane potential, $v_{mem}$, which integrates incoming spikes and generates an output spike whenever the membrane potential cross a particular threshold $v_{th}$. The membrane potential is reset to zero at the generation of an output spike. 
All neurons are reset whenever a spike train corresponding to a new image/pattern in presented. 
The IF Spiking Neuron dynamics as a function of time-step, $t$, can be described by the following equation,
\begin{equation}
v_{mem}(t+1) = v_{mem}(t) + \sum_{i} w_i.\mathbb{X}_i(t)
\label{lif_a}
\end{equation}
Note that the neuron dynamics is independent of the actual magnitude of the time-step. Let us first consider the simple case of a neuron being driven by a single input $\mathbb{X}(t)$ and a positive synaptic weight $w$. Due to the absence of any leak term in the neural dynamics, it is intuitive to show that the corresponding output spiking rate of the neuron is given by $\mathbb{E}[\mathbb{Y}(t)] \propto \mathbb{E}[\mathbb{X}(t)]$, with the proportionality factor being dependent on the ratio of $w$ and $v_{th}$. 
In the case when the synaptic weight is negative, the output spiking activity of the IF neuron is zero since the neuron is never able to cross the firing potential $v_{th}$, mirroring the functionality of a ReLU. 
The higher the ratio of the threshold with respect to the weight, the more time is required for the neuron to spike, thereby reducing the neuron spiking rate, $\mathbb{E}[\mathbb{Y}(t)]$, or equivalently increasing the time-delay for the neuron to generate a spike. A relatively high firing threshold can cause a huge delay for neurons to generate output spikes. 
For deep architectures, such a delay can quickly accumulate and cause the network to not produce any spiking outputs for relatively long periods of time. On the other hand, a relatively low threshold causes the SNN to lose any ability to distinguish between different magnitudes of the spike inputs being accumulated to the membrane potential (the term $\sum_{i} w_i.\mathbb{X}_i(t)$ in Eq. \ref{lif_a}) of the Spiking Neuron, causing it to lose evidence during the membrane potential integration process. This, in turn, results in accuracy degradation of the converted network. Hence, an appropriate choice of the ratio of the neuron threshold to the synaptic weights is essential to ensure minimal loss in classification accuracy during the ANN-SNN conversion process (\cite{diehl2015fast}). Consequently, most of the research work in this field has been concentrated on outlining appropriate algorithms for threshold-balancing, or equivalently, weight normalizing different layers of a network to achieve near-lossless ANN-SNN conversion.

\subsection{Architectural Constraints}
\subsubsection{Bias in Neural Units}
Typically neural units used for ANN-SNN conversion schemes are trained without any bias term (\cite{diehl2015fast}). This is due to the fact that optimization of the bias term in addition to the spiking neuron threshold expands the parameter space exploration, thereby causing the ANN-SNN conversion process to be more difficult. 
Requirement of bias less neural units also entails that Batch Normalization technique (\cite{ioffe2015batch}) cannot be used as a regularizer during the training process since it biases the inputs to each layer of the network to ensure each layer is provided with inputs having zero mean. Instead, we use dropout (\cite{srivastava2014dropout}) as the regularization technique. This technique simply masks portions of the input to each layer by utilizing samples from a Bernoulli distribution where each input to the layer has a specified probability of being dropped.

\subsubsection{Pooling Operation}
Deep convolutional neural network architectures typically consist of intermediate pooling layers to reduce the size of the convolution output maps. While various choices exist for performing the pooling mechanism, the two popular choices are either max-pooling (maximum neuron output over the pooling window) or spatial-averaging (two-dimensional average pooling operation over the pooling window). Since the neuron activations are binary in SNNs instead of analog values, performing max-pooling would result in significant information loss for the next layer. Consequently, we consider spatial-averaging as the pooling mechanism in this work (\cite{diehl2015fast}).

\section{Deep Convolutional SNN Architectures: VGG}
As mentioned previously, our work is based on the proposal outlined by authors in Ref. (\cite{diehl2015fast}) wherein the neuron threshold of a particular layer is set equal to the maximum activation of all ReLUs in the corresponding layer (by passing the entire training set through the trained ANN once after training is completed). Such a ``Data-Based Normalization" technique was evaluated for  three-layered fully connected and convolutional architectures on the MNIST dataset (\cite{diehl2015fast}). Note that, this process is referred to as ``weight-normalization" and ``threshold-balancing" interchangeably in this text. As mentioned before, the goal of this work is to optimize the ratio of the synaptic weights with respect to the neuron firing threshold, $v_{th}$. 
Hence, either all the synaptic weights preceding a neural layer are scaled by a normalization factor $w_{norm}$ equal to the maximum neural activation and the threshold is set equal to $1$ (``weight-normalization"), or the threshold $v_{th}$ is set equal to the maximum neuron activation for the corresponding layer with the synaptic weights remaining unchanged (``threshold-balancing"). Both operations are exactly equivalent mathematically.

\subsection{Proposed Algorithm: \textsc{Spike}-\textsc{Norm}}
However, the above algorithm leads us to the question: \textbf{Are ANN activations representative of SNN activations?} Let us consider a particular example for the case of maximum activation for a single ReLU. The neuron receives two inputs, namely $0.5$ and $1$. Let us consider unity synaptic weights in this scenario. Since the maximum ReLU activation is $1.5$, the neuron threshold would be set equal to $1.5$. However, when this network is converted to the SNN mode, both the inputs would be propagating binary spike signals. The ANN input, equal to $1$, would be converted to spikes transmitting at every time-step while the other input would transmit spikes approximately $50\%$ of the duration of a large enough time-window. Hence, the actual summation of spike inputs received by the neuron per time-step would be $2$ for a large number of samples, which is higher than the spiking threshold ($1.5$). Clearly, some information loss would take place due to the lack of this evidence integration.

Driven by this observation, we propose a weight-normalization technique that balances the threshold of each layer by considering the actual operation of the SNN in the loop during the ANN-SNN conversion process. The algorithm normalizes the weights of the network sequentially for each layer. Given a particular trained ANN, the first step is to generate the input Poisson spike train for the network over the training set for a large enough time-window. The Poisson spike train allows us to record the maximum summation of weighted spike-input (the term $\sum_{i} w_i.\mathbb{X}_i(t)$ in Eq. \ref{lif_a}, and hereafter referred to maximum SNN activation in this text) that would be received by the first neural layer of the network. In order to minimize the temporal delay of the neuron and simultaneously ensure that the neuron firing threshold is not too low, we weight-normalize the first layer depending on the maximum spike-based input received by the first layer. After the threshold of the first layer is set, we are provided with a representative spike train at the output of the first layer which enables us to generate the input spike-stream for the next layer. The process is continued sequentially for all the layers of the network. The main difference between our proposal and prior work (\cite{diehl2015fast}) is the fact that the proposed weight-normalization scheme accounts for the actual SNN operation during the conversion process. As we will show in the Results section, this scheme is crucial to ensure near-lossless ANN-SNN conversion for significantly deep architectures and for complex recognition problems. We evaluate our proposal for VGG-16 network (\cite{simonyan2014very}), a standard deep convolutional network architecture which consists of a 16 layer deep network composed of $3\times 3$ convolution filters (with intermediate pooling layers to reduce the output map dimensionality with increasing number of maps). The pseudo-code of the algorithm is given below.
\begin{algorithm}
\SetKwData{Left}{left}\SetKwData{This}{this}\SetKwData{Up}{up}
\SetKwFunction{Union}{Union}\SetKwFunction{FindCompress}{FindCompress}
\SetKwInOut{Input}{input}\SetKwInOut{Output}{output}

\Input{Input Poisson Spike Train $spikes$, Number of Time-Steps $\#timesteps$}
\Output{Weight-normalization / Threshold-balancing factors $v_{th,norm}[i]$ for each neural layer ($net.layer[i]$) of the network $net$}
initialization $v_{th,norm}[i]=0$ $\forall$ $i = 1,..., \#net.layer$\;
\tcp{Set input of 1st layer equal to spike train}
$net.layer[1].input$ = $spikes$\;
\For{$i\leftarrow 1$ \KwTo $\#net.layer$}{
\For{$t\leftarrow 1$ \KwTo $\#timesteps$}{
\tcp{Forward pass spike-train for neuron layer-i characterized by membrane potential $net.layer[i].v_{mem}$ and threshold $net.layer[i].v_{th}$}
$net.layer[i]:forward(net.layer[i].input[t])$ \;
\tcp{Determine threshold-balancing factor according to maximum SNN activation, max($net.layer[i].weight*net.layer[i].input[t]$), where '*' represents the dot-product operation}
$v_{th,norm}[i]$ = max($v_{th,norm}[i]$,max($net.layer[i].weight*net.layer[i].input[t]$))\;
}
\tcp{Threshold-balance layer-i}
$net.layer[i].v_{th}$ = $v_{th,norm}[i]$\;
\tcp{Record input spike-train for next layer}
$net.layer[i+1].input$ = $net.layer[i]:forward(net.layer[i].input)$\;
}
\caption{\textsc{Spike}-\textsc{Norm}}
\end{algorithm}

\section{Extension to Residual Architectures} 
\begin{figure}[!t]
\centering
\includegraphics[width=5.3in]{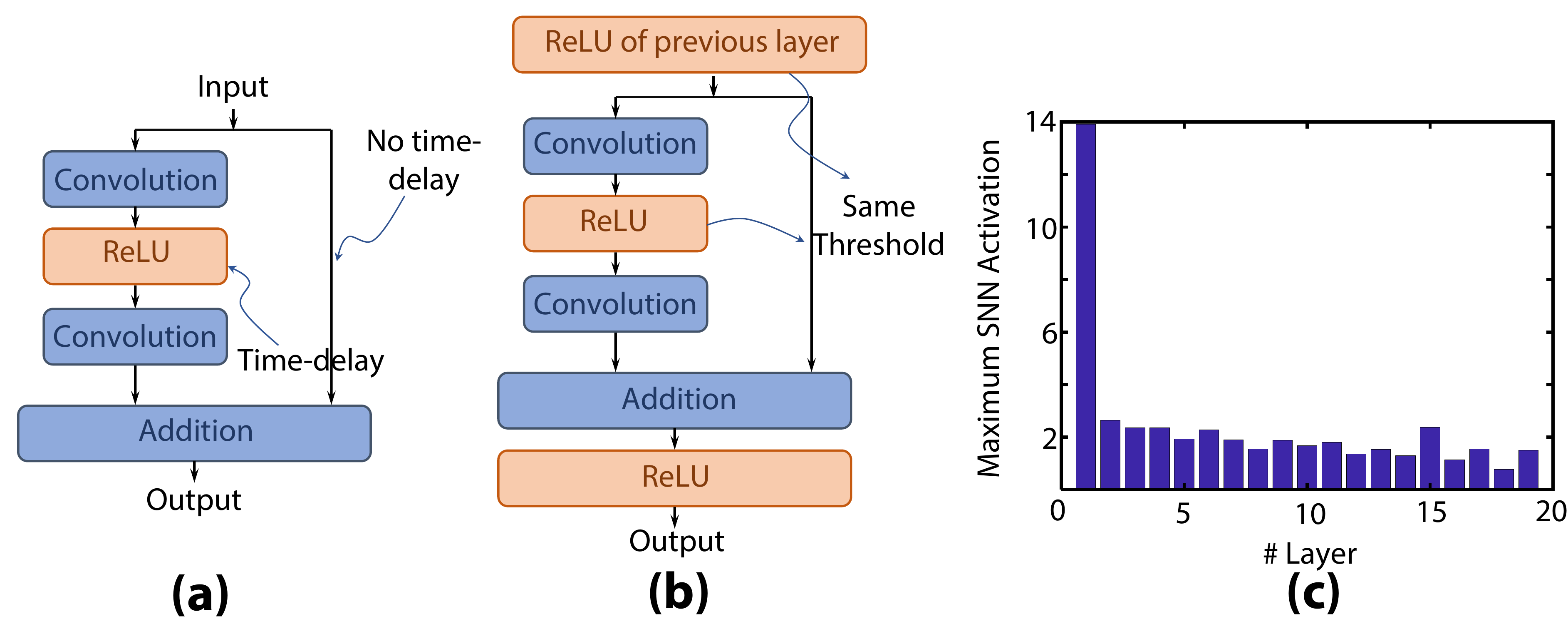}
\caption{(a) The basic ResNet functional unit, (b) Design constraints introduced in the functional unit to ensure near-lossless ANN-SNN conversion, (c) Typical maximum SNN activations for a ResNet having junction ReLU layers but the non-identity and identity input paths not having the same spiking threshold. While this is not representative of the case with equal thresholds in the two paths, it does justify the claim that after a few initial layers, the maximum SNN activations decay to values close to unity due to the identity mapping.}
\label{fig3_a}
\end{figure}
Residual network architectures were proposed as an attempt to scale convolutional neural networks to very deep layered stacks (\cite{he2016deep}). Although different variants of the basic functional unit have been explored, we will only consider identity shortcut connections in this text (shortcut type-A according to the paper (\cite{he2016deep})). Each unit consists of two parallel paths. The non-identity path consists of two spatial convolution layers with an intermediate ReLU layer. While the original ResNet formulation considers ReLUs at the junction of the parallel non-identity and identity paths (\cite{he2016deep}), recent formulations do not consider junction ReLUs in the network architecture (\cite{he2016identity}). Absence of ReLUs at the junction point of the non-identity and identity paths was observed to produce a slight improvement in classification accuracy on the CIFAR-10 dataset\footnote{http://torch.ch/blog/2016/02/04/resnets.html}. Due to the presence of the shortcut connections, important design considerations need to be accounted for to ensure near-lossless ANN-SNN conversion. We start with the basic unit, as shown in Fig. \ref{fig3_a}(a), and point-wise impose various architectural constraints with justifications. Note the discussion in this section is based on threshold-balancing (with synaptic weights remaining unscaled), i.e. the threshold of the neurons are adjusted to minimize ANN-SNN conversion loss.

\subsection{ReLUs at each junction point}
As we will show in the Results section, application of our proposed \textsc{Spike}-\textsc{Norm} algorithm on such a residual architecture resulted in a converted SNN that exhibited accuracy degradation in comparison to the original trained ANN. We hypothesize that this degradation is attributed mainly to the absence of any ReLUs at the junction points. Each ReLU when converted to an IF Spiking Neuron imposes a particular amount of characteristic temporal delay (time interval between an incoming spike and the outgoing spike due to evidence integration). Due to the shortcut connections, spike information from the initial layers gets instantaneously propagated to later layers. The unbalanced temporal delay in the two parallel paths of the network can result in distortion of the spike information being propagated through the network. Consequently, as shown in Fig. \ref{fig3_a}(b), we include ReLUs at each junction point to provide a temporal balancing effect to the parallel paths (when converted to IF Spiking Neurons). An ideal solution would be to include a ReLU in the parallel path, but that would destroy the advantage of the identity mapping.
\subsection{Same threshold of all fan-in layers}
As shown in the next section, direct application of our proposed threshold-balancing scheme still resulted in some amount of accuracy loss in comparison to the baseline ANN accuracy. 
However, note that the junction neuron layer receives inputs from the previous junction neuron layer as well as the non-identity neuron path. Since the output spiking activity of a particular neuron is also dependent on the threshold-balancing factor, all the fan-in neuron layers should be threshold-balanced by the same amount to ensure that input spike information to the next layer is rate-encoded appropriately. However, the spiking threshold of the neuron layer in the non-identity path is dependent on the activity of the neuron layer at the previous junction. An observation of the typical threshold-balancing factors for the network without using this constraint (shown in Fig. \ref{fig3_a}(c)) reveal that the threshold-balancing factors mostly lie around unity after a few initial layers. This occurs mainly due to the identity mapping. The maximum summation of spike inputs received by the neurons in the junction layers are dominated by the identity mapping (close to unity). From this observation, we heuristically choose both the thresholds of the non-identity ReLU layer and the identity-ReLU layer equal to $1$. However, the accuracy is still unable to approach the baseline ANN accuracy, which leads us to the third design constraint.
\subsection{Initial Non-Residual Pre-Processing Layers}
An observation of Fig. \ref{fig3_a}(c) reveals that the threshold-balancing factors of the initial junction neuron layers are significantly higher than unity. This can be a primary reason for the degradation in classification accuracy of the converted SNN. We note that the residual architectures used by authors in Ref. (\cite{he2016deep}) use an initial convolution layer with a very wide receptive field ($7 \times 7$ with a stride of $2$) on the ImageNet dataset. The main motive behind such an architecture was to show the impact of increasing depth in their residual architectures on the classification accuracy. Inspired by the VGG-architecture, we replace the first $7\times 7$ convolutional layer by a series of three $3 \times 3$ convolutions where the first two layers do not exhibit any shortcut connections. Addition of such initial non-residual pre-processing layers allows us to apply our proposed threshold-balancing scheme in the initial layers while using a unity threshold-balancing factor for the later residual layers. As shown in the Results section, this scheme significantly assists in achieving classification accuracies close to the baseline ANN accuracy since after the initial layers, the maximum neuron activations decay to values close to unity because of the identity mapping. 
\section{Experiments}
\subsection{Datasets and Implementation}
We evaluate our proposals on standard visual object recognition benchmarks, namely the CIFAR-10 and ImageNet datasets. Experiments performed on networks for the CIFAR-10 dataset are trained on the training set images with per-pixel mean subtracted and evaluated on the testing set. We also present results on the much more complex ImageNet 2012 dataset that contains 1.28 million training images and report evaluation (top-1 and top-5 error rates) on the $50,000$ validation set. $224 \times 224$ crops from the input images are used for this experiment.

We use VGG-16 architecture (\cite{simonyan2014very}) for both the datasets. ResNet-20 configuration outlined in Ref. (\cite{he2016deep}) is used for the CIFAR-10 dataset while ResNet-34 is used for experiments on the ImageNet dataset.  As mentioned previously, we do not utilize any batch-normalization layers. For VGG networks, a dropout layer is used after every ReLU layer except for those layers which are followed by a pooling layer. For Residual networks, we use dropout only for the ReLUs at the non-identity parallel paths but not at the junction layers. We found this crucial for achieving training convergence. Note that we have tested our framework only for the above mentioned architectures and datasets. There is no selection bias in the reported results.

Our implementation is derived from the Facebook ResNet implementation code for CIFAR and ImageNet datasets available publicly\footnotemark[\value{footnote}]. We use same image pre-processing steps and scale and aspect-ratio augmentation techniques as used in\footnotemark[\value{footnote}] (for instance, random crop, horizontal flip and color normalization transforms for the CIFAR-10 dataset). We report single-crop testing results while the error rates can be further reduced with 10-crop testing (\cite{krizhevsky2012imagenet}). Networks used for the CIFAR-10 dataset are trained on $2$ GPUs with a batchsize of $256$ for $200$ epochs, while ImageNet training is performed on $8$ GPUs for $100$ epochs with a similar batchsize. The initial learning rate is $0.05$. The learning rate is divided by $10$ twice, at $81$ and $122$ epochs for CIFAR-10 dataset and at $30$ and $60$ epochs for ImageNet dataset. A weight decay of $0.0001$ and a momentum of $0.9$ is used for all the experiments. Proper weight initialization is crucial to achieve convergence in such deep networks without batch-normalization. For a non-residual convolutional layer (for both VGG and ResNet architectures) having kernel size $k \times k$ with $n$ output channels, the weights are initialized from a normal distribution and standard deviation $\sqrt{\frac{2}{k^2n}}$. However, for residual convolutional layers, the standard deviation used for the normal distribution was $\frac{\sqrt{2}}{k^2n}$. We observed this to be important for achieving training convergence and a similar observation was also outlined in Ref. (\cite{hardt2016identity}) although their networks were trained without both dropout and batch-normalization.
\footnotetext{https://github.com/facebook/fb.resnet.torch}

\subsection{Experiments for VGG Architectures}
Our VGG-16 model architecture follows the implementation outlined in \footnote{https://github.com/szagoruyko/cifar.torch} except that we do not utilize the batch-normalization layers. We used a randomly chosen mini-batch of size 256 from the training set for the weight-normalization process on the CIFAR-10 dataset. While the entire training set can be used for the weight-normalization process, using a representative subset did not impact the results. We confirmed this by running multiple independent runs for both the CIFAR and ImageNet datasets. The standard deviation of the final classification error rate after $2500$ time-steps was $\sim 0.01\%$. 
All results reported in this section represent the average of 5 independent runs of the spiking network (since the input to the network is a random process). No notable difference in the classification error rate was observed at the end of $2500$ time-steps and the network outputs converged to deterministic values despite being driven by stochastic inputs. 
For the SNN model based weight-normalization scheme (\textsc{Spike}-\textsc{Norm} algorithm) we used $2500$ time-steps for each layer sequentially to normalize the weights. 

Table \ref{table_1_a} summarizes our results for the CIFAR-10 dataset. The baseline ANN error rate on the testing set was $8.3\%$. Since the main contribution of this work is to minimize the loss in accuracy during conversion from ANN to SNN for deep-layered networks and not in pushing state-of-the-art results in ANN training, we did not perform any hyper-parameter optimization. 
However, note that despite several architectural constraints being present in our ANN architecture, we are able to train deep networks that provide competitive classification accuracies using the training mechanisms described in the previous subsection. Further reduction in the baseline ANN error rate is possible by appropriately tuning the learning parameters. For the VGG-16 architecture, our implementation of the ANN-model based weight-normalization technique, proposed by Ref. (\cite{diehl2015fast}), yielded an average SNN error rate of $8.54\%$ leading to an error increment of $0.24\%$. The error increment was minimized to $0.15\%$ on applying our proposed \textsc{Spike}-\textsc{Norm} algorithm. Note that we consider a strict model-based weight-normalization scheme to isolate the impact of considering the effect of an ANN versus our SNN model for threshold-balancing. Further optimizations of considering the maximum synaptic weight during the weight-normalization process (\cite{diehl2015fast}) is still possible. 

Previous works have mainly focused on much shallower convolutional neural network architectures. Although Ref. (\cite{hunsberger2016training}) reports results with an accuracy loss of $0.18\%$, their baseline ANN suffers from some amount of accuracy degradation since their networks are trained with noise (in addition to architectural constraints mentioned before) to account for neuronal response variability due to incoming spike trains (\cite{hunsberger2016training}). It is also unclear whether the training mechanism with noise would scale up to deeper layered networks. \textbf{Our work reports the best performance of a Spiking Neural Network on the CIFAR-10 dataset till date.}

The impact of our proposed algorithm is much more apparent on the more complex ImageNet dataset. The rates for the top-1 (top-5) error on the ImageNet validation set are summarized in Table \ref{table_2_a}. Note that these are single-crop results. The accuracy loss during the ANN-SNN conversion process is minimized by a margin of $0.57\%$ by considering SNN-model based weight-normalization scheme. It is therefore expected that our proposed \textsc{Spike}-\textsc{Norm} algorithm would significantly perform better than an ANN-model based conversion scheme as the pattern recognition problem becomes more complex since it accounts for the actual SNN operation during the conversion process. Note that Ref. (\cite{hunsberger2016training}) reports a performance of $48.2\% (23.8\%)$ on the first 3072-image test batch of the ImageNet $2012$ dataset. 

At the time we developed this work, we were unaware of a parallel effort to scale up the performance of SNNs to deeper networks and large-scale machine learning tasks. The work was recently published in Ref. (\cite{rueckauer2017conversion}). However, their work differs from our approach in the following aspects: 
\newline (i) Their work improves on prior approach outlined in Ref. (\cite{diehl2015fast}) by proposing conversion methods for removing the constraints involved in ANN training (discussed in Section 4.3). We are improving on prior art by scaling up the methodology outlined in Ref. (\cite{diehl2015fast}) for ANN-SNN conversion by including the constraints.
\newline (ii) We are demonstrating that considering SNN operation in the conversion process helps to minimize the conversion loss. Ref. (\cite{rueckauer2017conversion}) uses ANN based normalization scheme used in Ref. (\cite{diehl2015fast}). 
\newline While removing the constraints in ANN training allows authors in Ref. (\cite{rueckauer2017conversion}) to train ANNs with better accuracy, they suffer significant accuracy loss in the conversion process. This occurs due to a non-optimal ratio of biases/batch-normalization factors and weights (\cite{rueckauer2017conversion}). This is the primary reason for our exploration of ANN-SNN conversion without bias and batch-normalization. For instance, their best performing network on CIFAR-10 dataset incurs a conversion loss of $1.06\%$ in contrast to $0.15\%$ reported by our proposal for a much deeper network. The accuracy loss is much larger for their VGG-16 network on the ImageNet dataset - $14.28\%$ in contrast to $0.56\%$ for our proposal. Although Ref. (\cite{rueckauer2017conversion}) reports a top-1 SNN error rate of $25.40\%$ for an Inception-V3 network, their ANN is trained with an error rate of $23.88\%$. The resulting conversion loss is $1.52\%$ and much higher than our proposals. The Inception-V3 network conversion was also optimized by a voltage clamping method, that was found to be specific for the Inception network and did not apply to the VGG network (\cite{rueckauer2017conversion}). Note that the results reported on ImageNet in Ref. (\cite{rueckauer2017conversion}) are on a subset of $1382$ image samples for Inception-V3 network and 2570 samples for VGG-16 network. Hence, the performance on the entire dataset is unclear. Our contribution lies in the fact that we are demonstrating ANNs can be trained with the above-mentioned constraints with competitive accuracies on large-scale tasks and converted to SNNs in a near-lossless manner. 

\textbf{This is the first work that reports competitive performance of a Spiking Neural Network on the entire $50,000$ ImageNet 2012 validation set.}

\begin{table}[t]
\renewcommand{\arraystretch}{1}
\small
\caption{Results for CIFAR-10 Dataset}
\label{table_1_a}
\centering
\begin{tabular}{ p{6cm} p{2.1cm} p{2.1cm} p{3.4cm} }
\hline 
\hline
\bfseries {Network Architecture} & \bfseries {ANN \newline Error} & \bfseries {SNN \newline Error} & \bfseries {Error Increment}\\
\hline
{4-layered networks (\cite{cao2015spiking}) \newline (Input cropped to 24 x 24)} & {$20.88\%$} & {$22.57\%$} & {$1.69\%$}\\ \\
{3-layered networks (\cite{esser2016convolutional})} & {$-$} & {$10.68\%$} & {$-$}\\ \\
{8-layered networks (\cite{hunsberger2016training}) \newline (Input cropped to 24 x 24)} & {$16.28\%$} & {$16.46\%$} & {0.18\%}\\ \\
{6-layered networks (\cite{rueckauer2017conversion}) \newline } & {$8.09\%$} & {$9.15\%$} & {1.06\%}\\ \\
{VGG-16 \newline(ANN model based conversion)} & {$8.3\%$} & {$8.54\%$} & {$0.24\%$}\\ \\
{\textbf{VGG-16 \newline(\textsc{SPIKE}-\textsc{NORM})}} & {\textbf{8.3\%}} & {\textbf{8.45\%}} & {\textbf{0.15\%}}\\
\hline
\hline
\end{tabular}
\end{table}

\subsection{Experiments for Residual Architectures}

\begin{figure}[!t]
\centering
\includegraphics[width=2.3in]{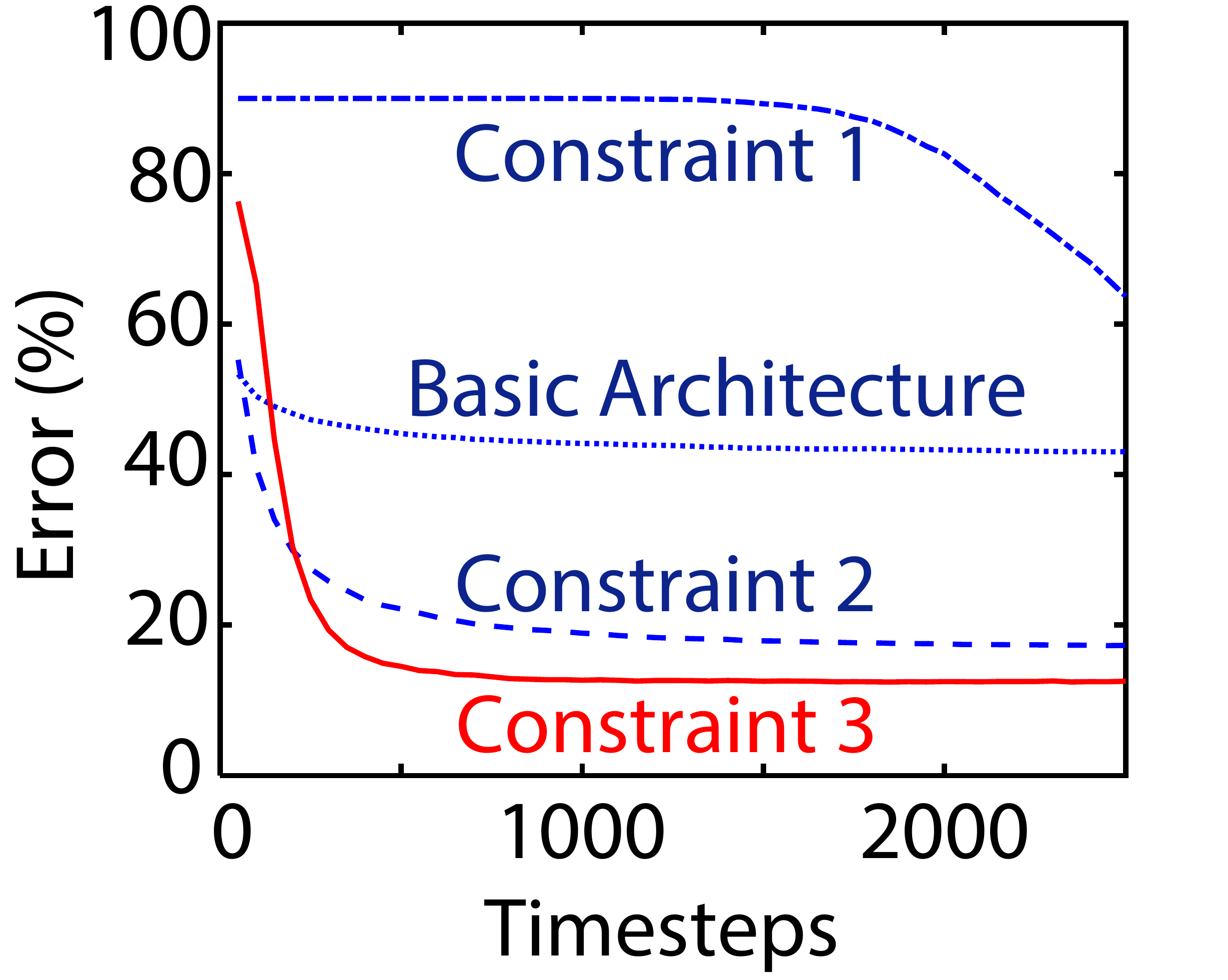}
\caption{Impact of the architectural constraints for Residual Networks. ``Basic Architecture" does not involve any junction ReLU layers. ``Constraint 1" involves junction ReLUs while ``Constraint 2" imposes equal unity threshold for all residual units. Network accuracy is significantly improved with the inclusion of ``Constraint 3" that involves pre-processing weight-normalized plain convolutional layers at the network input stage. }
\label{fig4_a}
\end{figure}

Our residual networks for CIFAR-10 and ImageNet datasets follow the implementation in Ref. (\cite{he2016deep}). We first attempt to explain our design choices for ResNets by sequentially imposing each constraint on the network and showing their corresponding impact on network performance in Fig. \ref{fig4_a}. The ``Basic Architecture" involves a residual network without any junction ReLUs. ``Constraint 1" involves junction ReLUs without having equal spiking thresholds for all fan-in neural layers. ``Constraint 2" imposes an equal threshold of unity for all the layers while ``Constraint 3" performs best with two pre-processing plain convolutional layers ($3 \times 3$) at the beginning of the network. The baseline ANN ResNet-20 was trained with an error of $10.9\%$ on the CIFAR-10 dataset. Note that although we are using terminology consistent with Ref. (\cite{he2016deep}) for the network architectures, our ResNets contain two extra plain pre-processing layers. The converted SNN according to our proposal yielded a classification error rate of $12.54\%$. Weight-normalizing the initial two layers using the ANN-model based weight-normalization scheme produced an average error of $12.87\%$, further validating the efficiency of our weight-normalization technique. 

\begin{figure}[!t]
\centering
\includegraphics[width=3.8in]{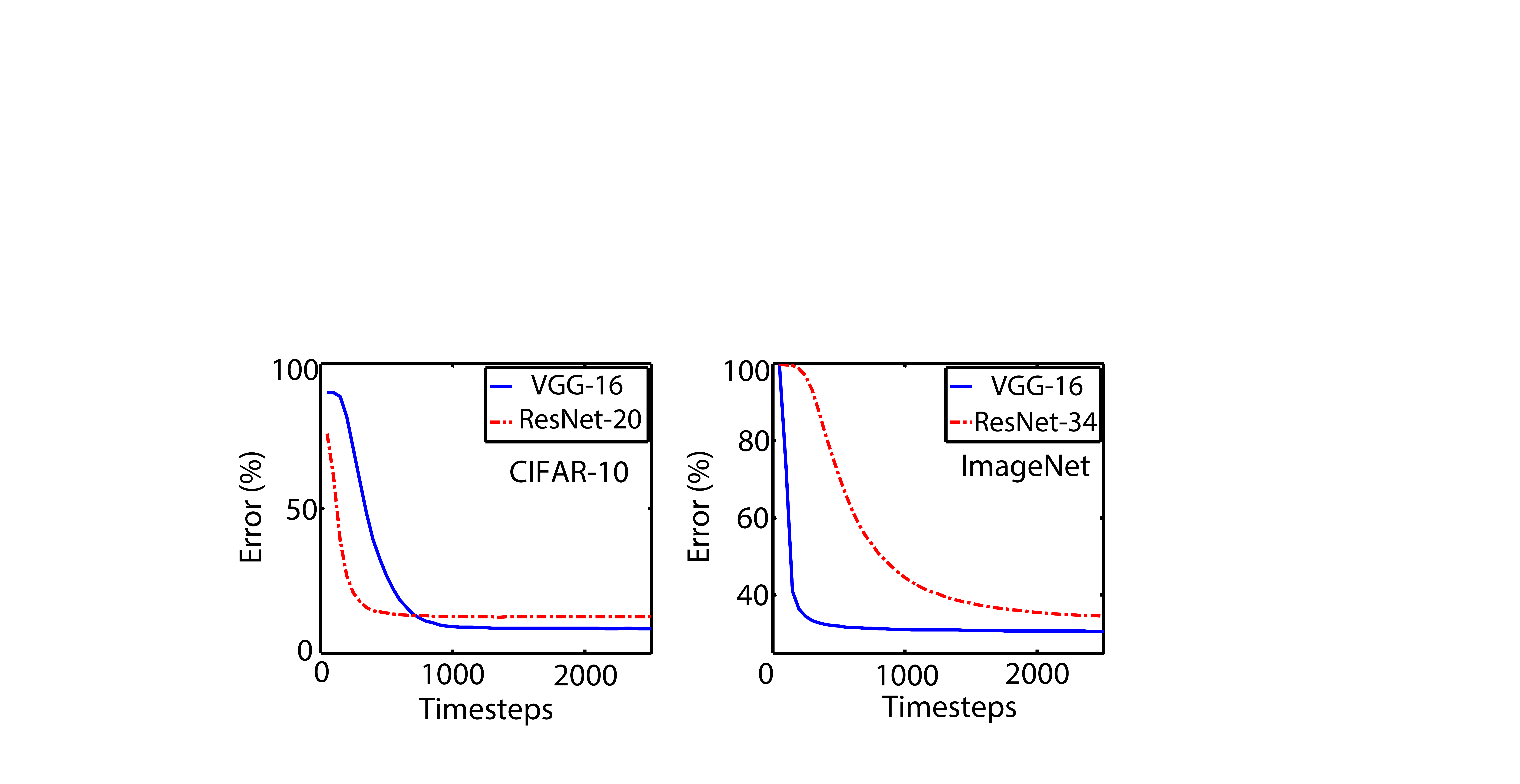}
\caption{Convergence plots for the VGG and ResNet SNN architectures for CIFAR-10 and ImageNet datasets are shown above. The classification error reduces as more evidence is integrated in the Spiking Neurons with increasing time-steps. Note that although the network depths are similar for CIFAR-10 dataset, the ResNet-20 converges much faster than the VGG architecture. The delay for inferencing is higher for ResNet-34 on the ImageNet dataset due to twice the number of layers as the VGG network.}
\label{fig5_a}
\end{figure}

On the ImageNet dataset, we use the deeper ResNet-34 model outlined in Ref. (\cite{he2016deep}). The initial $7 \times 7$ convolutional layer is replaced by three $3 \times 3$ convolutional layers where the initial two layers are non-residual plain units. The baseline ANN is trained with an error of $29.31\%$ while the converted SNN error is $34.53\%$ at the end of $2500$ timesteps. The results are summarized in Table. \ref{table_3_a} and convergence plots for all our networks are provided in Fig. \ref{fig5_a}.
\begin{table}[t]
\small
\renewcommand{\arraystretch}{1.3}
\caption{Results for ImageNet Dataset}
\label{table_2_a}
\centering
\begin{tabular}{ p{6.4cm} p{2.1cm} p{2.1cm} p{3.4cm} }
\hline 
\hline
\bfseries {Network Architecture} & \bfseries {ANN \newline Error} & \bfseries {SNN \newline Error} & \bfseries {Error Increment}\\
\hline
{8-layered networks (\cite{hunsberger2016training}) \newline(Tested on subset of 3072 images)} & {$-$} & {$48.20\% \newline (23.80\%)$} & {$-$}\\ \\
{VGG-16 (\cite{rueckauer2017conversion}) \newline(Tested on subset of 2570 images)} & {$36.11\% \newline (15.14\%)$} & {$50.39\% \newline (18.37\%)$} & {$14.28\% \newline (3.23\%)$}\\ \\
{VGG-16 \newline(ANN model based conversion)} & {$29.48\% \newline (10.61\%)$} & {$30.61\% \newline (11.21\%)$} & {$1.13\% \newline (0.6\%)$}\\ \\
{\textbf{VGG-16 \newline(\textsc{SPIKE}-\textsc{NORM})}} & {\textbf{29.48\% \newline (10.61\%)}} & {\textbf{30.04\% \newline (10.99\%)}} & {\textbf{0.56\% \newline (0.38\%)}}\\
\hline
\hline
\end{tabular}

\end{table}
It is worth noting here that the main motivation of exploring Residual Networks is to go deeper in Spiking Neural Networks. We explore relatively simple ResNet architectures, as the ones used in Ref. (\cite{he2016deep}), which have an order of magnitude fewer parameters than standard VGG-architectures. Further hyper-parameter optimizations or more complex architectures are still possible. While the accuracy loss in the ANN-SNN conversion process is more for ResNets than plain convolutional architectures, yet further optimizations like including more pre-processing initial layers or better threshold-balancing schemes for the residual units can still be explored. \textbf{This work serves as the first work to explore ANN-SNN conversion schemes for Residual Networks and attempts to highlight important design constraints required for minimal loss in the conversion process.}
\subsection{Computation Reduction Due to Sparse Neural Events}
\begin{figure}[!t]
\centering
\includegraphics[width=2.5in]{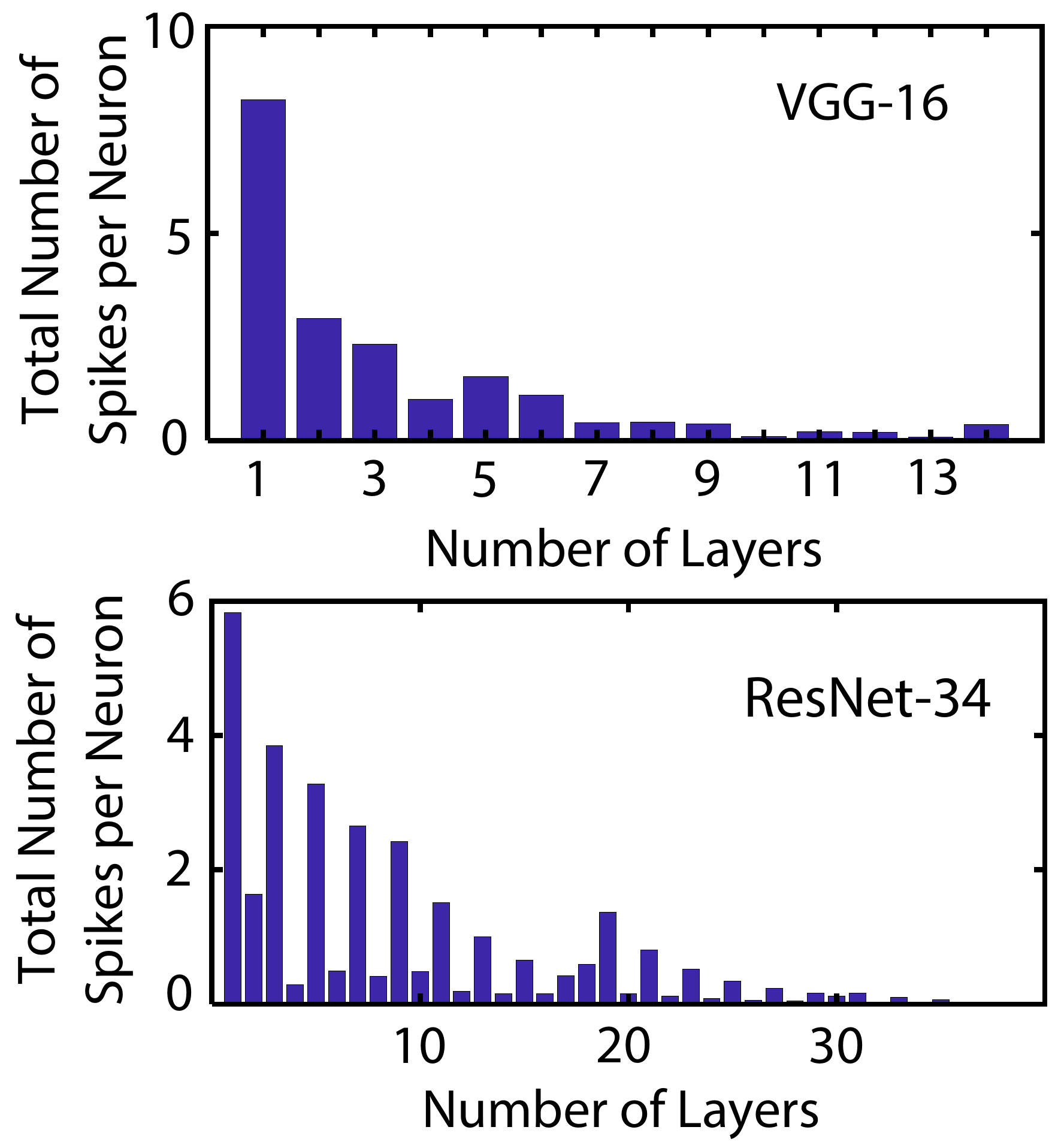}
\caption{Average cumulative spike count generated by neurons in VGG and ResNet architectures on the ImageNet dataset as a function of the layer number. $500$ timesteps were used for accumulating the spike-counts for VGG networks while $2000$ time-steps were used for ResNet architectures. The neural spiking sparsity increases significantly as network depth increases.}
\label{fig6_a}
\end{figure}
ANN operation for prediction of the output class of a particular input requires a single feed-forward pass per image. For SNN operation, the network has to be evaluated over a number of time-steps. However, specialized hardware that accounts for the event-driven neural operation and ``computes only when required" can potentially exploit such alternative mechanisms of network operation. For instance, Fig. \ref{fig6_a} represents the average total number of output spikes produced by neurons in VGG and ResNet architectures as a function of the layer for ImageNet dataset. A randomly chosen minibatch was used for the averaging process. We used $500$ timesteps for accumulating the spike-counts for VGG networks while $2000$ time-steps were used for ResNet architectures. This is in accordance to the convergence plots shown in Fig. \ref{fig5_a}. An important insight obtained from Fig. \ref{fig6_a} is the fact that neuron spiking activity becomes sparser as the network depth increases. Hence, benefits from event-driven hardware is expected to increase as the network depth increases. While an estimate of the actual energy consumption reduction for SNN mode of operation is outside the scope of this current work, we provide an intuitive insight by providing the number of computations per synaptic operation being performed in the ANN versus the SNN. 

The number of synaptic operations per layer of the network can be easily estimated for an ANN from the architecture for the convolutional and linear layers. For the ANN, a multiply-accumulate (MAC) computation takes place per synaptic operation. On the other hand, a specialized SNN hardware would perform an accumulate computation (AC) per synaptic operation only upon the receipt of an incoming spike. Hence, the total number of AC operations occurring in the SNN would be represented by the dot-product of the average cumulative neural spike count for a particular layer and the corresponding number of synaptic operations. Calculation of this metric reveal that for the VGG network, the ratio of SNN AC operations to ANN MAC operations is $1.975$ while the ratio is $2.4$ for the ResNet (the metric includes only ReLU/IF spiking neuron activations in the network). However, note the fact that a MAC operation involves an order of magnitude more energy consumption than an AC operation (\cite{han2015learning}). Hence, the energy consumption reduction for our SNN implementation is expected to be significantly lower in comparison to the original ANN implementation. It is worth noting here that the real metric governing the energy requirement of SNN versus ANN is the number of spikes per neuron. Energy benefits are obtained only when the average number of spikes per neuron over the inference timing window is less than 1 (since in the SNN the synaptic operation is conditional based on spike generation by neurons in the previous layer). Hence, to get benefits for energy reductions in SNNs, one should target deeper networks due to neuron spiking sparsity. 

While the SNN operation requires a number of time-steps in contrast to a single feed-forward pass for an ANN, the actual time required to implement a single time-step of the SNN in a neuromorphic architecture might be significantly lower than a feedforward pass for an ANN implementation (due to event-driven hardware operation). An exact estimate of the delay comparison is outside the scope of this article. Nevertheless, despite the delay overhead, as highlighted above, the power benefits of event-driven SNNs can significantly increase the energy (power x delay) efficiency of deep SNNs in contrast to ANNs.

\begin{table}[t]
\renewcommand{\arraystretch}{1.3}
\small
\caption{Results for Residual Networks}
\label{table_3_a}
\centering
\begin{tabular}{ p{2.4cm} p{2.8cm} p{2.4cm} p{2.4cm} }
\hline 
\hline
\bfseries {Dataset} & \bfseries {Network \newline Architecture} & \bfseries {ANN \newline Error} & \bfseries {SNN \newline Error}\\
\hline
{CIFAR-10} & {ResNet-20} & {$10.9\%$} & {$12.54\%$}\\
{ImageNet} & {ResNet-34} & {$29.31\% \newline (10.31\%)$} & {$34.53\% \newline (13.67\%)$}\\
\hline
\hline
\end{tabular}
\end{table}
\section{Conclusions and Future Work}
This work serves to provide evidence for the fact that SNNs exhibit similar computing power as their ANN counterparts. This can potentially pave the way for the usage of SNNs in large scale visual recognition tasks, which can be enabled by low-power neuromorphic hardware. However, there are still open areas of exploration for improving SNN performance. A significant contribution to the present success of deep NNs is attributed to Batch-Normalization (\cite{ioffe2015batch}). While using bias less neural units constrain us to train networks without Batch-Normalization, algorithmic techniques to implement Spiking Neurons with a bias term should be explored. Further, it is desirable to train ANNs and convert to SNNs without any accuracy loss. Although the proposed conversion technique attempts to minimize the conversion loss to a large extent, yet other variants of neural functionalities apart from ReLU-IF Spiking Neurons could be potentially explored to further reduce this gap. Additionally, further optimizations to minimize the accuracy loss in ANN-SNN conversion for ResNet architectures should be explored to scale SNN performance to even deeper architectures.

\bibliographystyle{frontiersinSCNS_ENG_HUMS} 

\end{document}